# An End-to-end Supervised Domain Adaptation Framework for Cross-Domain Change Detection

**Jia Liu**[a,c,†], **Wenjie Xuan**[b,c,d,e,†], **Yuhang Gan**[b,c,d,e,f], **Yibing Zhan**[g], **Juhua Liu**[a,c*], **Bo Du**[b,c,d,e]

[a] Research Center for Graphic Communication, Printing and Packaging, Wuhan University, Wuhan, China.

[b] National Engineering Research Center for Multimedia Software, Wuhan University, Wuhan, China.

[c] Institute of Artificial Intelligence, Wuhan University, Wuhan, China.

[d] School of Computer Science, Wuhan University, Wuhan, China.

[e] Hubei Key Laboratory of Multimedia and Network Communication Engineering, Wuhan University, Wuhan, China.

[f] Land Satellite Remote Sensing Application Center, MNR, Beijing, China.

[g] JD Explore Academy, Beijing, China.

**Abstract.** Change detection is a crucial but extremely challenging task in remote sensing image analysis, and much progress has been made with the rapid development of deep learning. However, most existing deep learning-based change detection methods try to elaborately design complicated neural networks with powerful feature representations. However, they ignore the universal domain shift induced by time-varying land cover changes, including luminance fluctuations and seasonal changes between pre-event and post-event images, thereby producing suboptimal results. In this paper, we propose an end-to-end supervised domain adaptation framework for cross-domain change detection named SDACD, to effectively alleviate the domain shift between bi-temporal images for better change predictions. Specifically, our SDACD presents collaborative adaptations from both image and feature perspectives with supervised learning. Image adaptation exploits generative adversarial learning with cycle-consistency constraints to perform cross-domain style transformation, which effectively narrows the domain gap in a two-side generation fashion. As for feature adaptation, we extract domain-invariant features to align different feature distributions in the feature space, which could further reduce the domain gap of cross-domain images. To further improve the performance, we combine three types of bi-temporal images for the final change prediction, including the initial input bi-temporal images and two generated bi-temporal images from the pre-event and post-event domains. Extensive experiments and analyses conducted on two benchmarks demonstrate the effectiveness and generalizability of our proposed framework. Notably, our framework pushes several representative baseline models up to new State-Of-The-Art records, achieving 97.34% and 92.36% on the CDD and WHU building datasets, respectively. The source code and models are publicly available at https://github.com/Perfect-You/SDACD.

[†] **Jia Liu and Wenjie Xuan contributed equally to this work.**

**Address all correspondence to:**

* Juhua Liu, Research Center for Graphic Communication, Printing and Packaging, Wuhan University, Wuhan, China; Tel: +86-18062452253; E-mail: liujuhua@whu.edu.cn.

# 1 Introduction

Change detection (CD) aims to identify significant differences in geographical elements between bi-temporal images of the same geographic area. It takes registered bi-temporal images as the input and outputs pixel-wise change maps. This fundamental but important remote sensing task has gradually become an active topic in the computer vision community due to its wide applications in urbanization monitoring [1], resource and environment monitoring [2], disaster assessment [3], etc. Many excellent methods have been proposed recently due to the easy acquisition of high-resolution remote sensing images and the success of deep learning. However, many issues in this task remain open and challenging due to the complex and heterogeneous appearance of geographical elements at different times.

During the past decades, numerous kinds of conventional change detection methods have been proposed, which can be classified into four categories: 1) **Algebra-based methods** perform channel-wise algebraic operations on the registered bi-temporal images directly, including image differencing, image regression, and change vector analysis [4]. However, it is scene-dependent and time-consuming to find a suitable threshold to distinguish changed pixels from unchanged ones. 2) **Transformation-based methods** cast bi-temporal images into specific feature spaces capable of narrowing the pixel differences of unchanged regions and highlighting the changed information. However, methods such as [5][6] struggle to handle high-resolution images because they depend on empirically designed features. 3) **Classification-based methods** such as [7][8] identify change regions by comparing the pre-generated land cover labels of geographic elements between bi-temporal images. However, the classification errors of the pre-generated labels would accumulate on the final change maps and inevitably reduce the accuracy of change predictions. 4) **Machine learning-based methods** employ traditional machine learning algorithms, such as random forest regression [9] and support vector machine [10], to determine whether the specified area has changed. Although most of these conventional methods are simple and explainable, they show poor robustness in real scenarios because their performance is sensitive to noise and limited by handcrafted features.

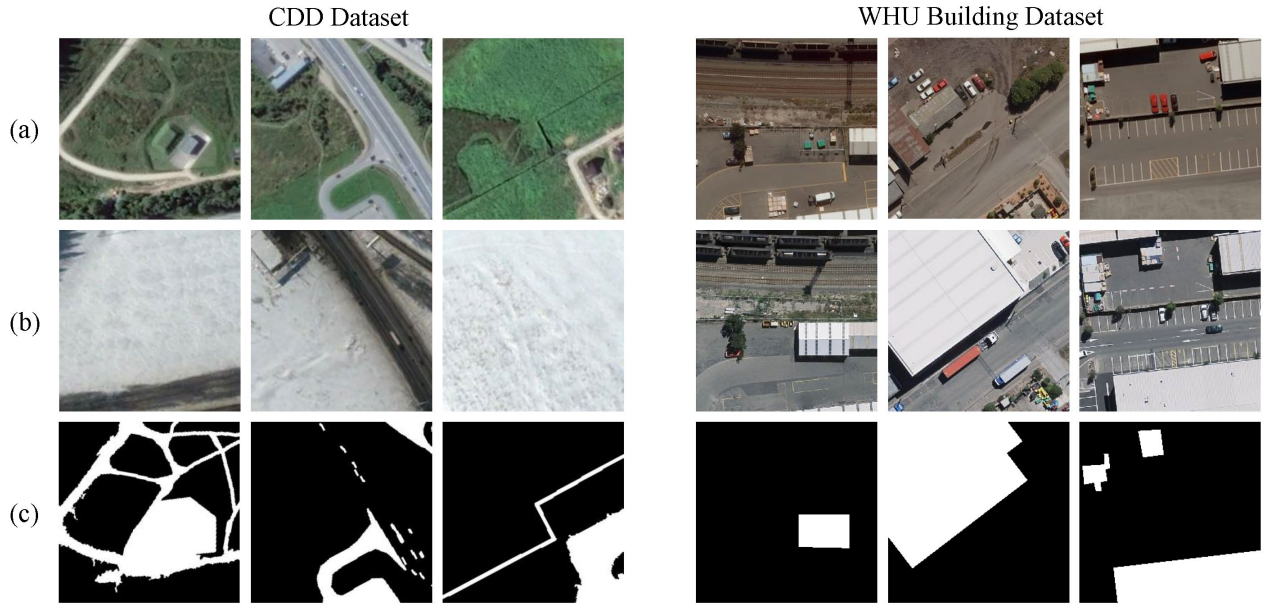

**Fig. 1** The cross-domain bi-temporal images in the CDD and WHU building datasets. Images from top to bottom are (a) pre-event images, (b) post-event images, and (c) corresponding ground truths, respectively. The bi-temporal images of the CDD dataset were acquired in summer and winter, while the WHU building dataset was taken under varied luminance conditions in 2012 and 2016, respectively. Both datasets exhibit significant appearance variations.

Recently, with the rapid development of deep learning, great progress has been made in the computer vision community [11]-[15], and many deep learning-based change detection methods have also been proposed. Due to the powerful ability of automatic high-level feature extraction, they have demonstrated superior performance and robustness to conventional methods. Deep learning-based methods usually first extract discriminative features from bi-temporal images using convolutional neural networks similar to Siamese networks, and then utilize fully convolutional networks, such as FC-Siam-diff [16] and SNUNet [17], or metric-based methods, such as STANet [18] and DASNet [19], to predict the final change results. For accurately predicting change results from the feature space, the features extracted from bi-temporal images should satisfy the following constraints: feature vectors associated with changed pixel pairs are farther apart from each other, while invariant pixel pairs are close. However, this assumption is difficult to satisfy in reality. Since bi-temporal images are usually acquired at different times, their imaging sensors, atmospheric and luminance conditions would be different. This inevitably leads to completely different appearances in the bi-temporal images regardless of whether the geographical elements have changed. As shown in Fig. 1, the time-varying changes introduced by seasons and climates are more obvious than the real ones caused by human factors, such as the construction or destruction of buildings and roads. These pseudo changes make it difficult for existing deep learning-based methods, which usually focus on designing complicated neural networks with powerful feature representation while ignoring the domain shift between bi-temporal images, to extract features satisfying the above constraints from the bi-temporal images.

In this paper, we propose a supervised domain adaptation framework for cross-domain change detection, named SDACD, to effectively alleviate the severe domain shift between the pre-event and post-event images. Our proposed SDACD presents collaborative adaptations from both image and feature perspectives and comprises two key modules: the image adaptation (IA) module and the feature adaptation (FA) module. The image adaptation module addresses domain shift by aligning the image appearance between bi-temporal domains via image-to-image transformation. Specifically, we transform the appearances of pre-event images and post-event images to each other by using bi-directional generative adversarial networks with cycle-consistency constraints, *i.e.*, we transform pre-event images to the appearance of post-event images and post-event images to pre-event images. Afterward, we employ two types of bi-temporal images from the pre-event domain and post-event domain as well as the original bi-temporal images to train the change detection model in the feature adaptation module, where we integrate feature adaptation into the framework to further narrow the remaining domain shift. Specifically, we first input three bi-temporal images, *i.e.*, the pre-event and post-event images, pre-event and post-event stylized as pre-event images, and post-event and pre-event stylized as post-event images, into the feature adaptation module and predict the change map for each pair of bi-temporal images. Then, we design a feature domain-invariance discriminator, which connects the change predictions and the bi-temporal images, to differentiate the predictions generated from the bi-temporal images. If this discriminator fails to distinguish, it means the extracted features for predicting change maps are domain-invariant. Finally, to fully utilize the image information from different domains, we make the final change prediction by fusing the above three features from different bi-temporal images. Our proposed framework combines the image and feature adaptation procedures in an end-to-end trainable manner, which makes them benefit from each other and achieve better performance. To the best of our knowledge, this is the first work that fully studies domain adaptation with supervised learning for change detection and provides a simple and effective solution that can be easily plugged in any non-domain adaptation model to further improve their performance. This is especially friendly for real-world applications. The architecture of our proposed framework is shown in Fig. 2.

We implement the SDACD framework based on three representative baseline models, STANet [16], DASNet [19], and SNUNet [17], and validate its effectiveness and generalizability on the CDD [21] and WHU building datasets [22]. Our framework shows consistent improvements on all three baseline models. For the CDD dataset, the *F1-scores* of the STANet, DASNet, and SNUNet-based frameworks improve by 0.70%, 0.78%, and 1.66%, respectively, while for the WHU building dataset, the scores improve by 8.81%, 2.59%, and 6.85%, respectively. Moreover, when combined with the cutting-edge SNUNet, our framework achieves state-of-the-art performance on both datasets. The visualized results also demonstrate the superiority of our framework in tackling bi-temporal change detection under severe domain shift, as illustrated in Section 4.

Our contributions can be summarized as follows:

- We propose a novel supervised domain adaptation framework SDACD for cross-domain change detection. It unifies image adaptation and feature adaptation in an end-to-end trainable manner to alleviate the domain shift between the pre-event and post-event images;

- Our framework is compatible with existing change detection networks that do not take into account domain shift. Furthermore, it can handle cross-domain change detection and consistently improve the performance as an easy-to-plug-in module;

- Experimental results on two benchmark datasets demonstrate the effectiveness and generalizability of our SDACD. Most importantly, our SNUNet-based framework sets new state-of-the-art performance with an *F1-score* of 97.34% on the CDD dataset and 92.36% on the WHU building dataset.

The rest of the paper is organized as follows. In Section 2, we briefly review the related works of change detection and domain adaptation. In Section 3, we introduce our proposed framework before analyzing each module. Section 4 reports and discusses the experimental results. Finally, we conclude in Section 5.

## 2 Related Works

### 2.1 Change detection

In recent years, due to the powerful feature extraction capabilities of deep neural networks, many deep learning-based change detection methods have been proposed. They can be divided into two main categories, namely, patch-based methods and image-based methods.

**Patch-based** methods [23]-[25] regard patches as the smallest processing unit for change detection. These methods first split bi-temporal images into patches and then decide whether each patch has changed or not. Zhang et al. [23] employed a deep belief network to transform multi-spectral image patches into specified feature spaces and relieve noise for change detection. To facilitate the utilization of spectral-spatial-temporal representations, Mou et al. [24] designed a ReCNN, which combined CNNs with RNNs to explore spectral-spatial representations and temporal dependencies between the patches of bi-temporal images. Although patch-based methods have made remarkable progress compared with traditional methods, they suffer from high memory consumption and low computational efficiency. Furthermore, patch size restricts the receptive field. Thus, these methods cannot exploit global information, which also greatly limits their performance.

**Image-based** methods [16]-[20] directly employ the whole image to make pixel-wise predictions. This is computationally efficient and leads to good performance. FC-EF, FC-Siam-conc, and FC-Siam-diff [16] are some earlier image-based methods. These methods employed UNet as the backbone network of feature extractors and explored three different feature fusion strategies, which greatly improved the performance and processing speed compared to patch-based methods. Later, Chen et al. [18] proposed an influential STANet, which designed a spatial-temporal self-attention module in the Siamese network to explore spatial-temporal dependencies between bi-temporal images. Recently, Fang et al. [17] proposed a densely connected SNUNet and achieved SOTA on the CDD dataset. SNUNet adopted a Siamese UNet++ [26] architecture as the feature extractor and utilized an ensemble channel attention module to aggregate multilevel features to enhance discriminative feature representations.

Although the aforementioned methods have achieved promising results by using powerful feature extractors and elaborately designed structures, they do not consider the universal domain shift between bi-temporal images caused by different seasons, various imaging conditions, etc. This produces suboptimal results. DLSF [27] and PDA [28] are recent representative methods for cross-domain change detection that employ GAN with complex constraints to preserve semantic information and bridge the style gaps between bi-temporal images. Our method differs from DLSF and PDA in two ways. First, these two methods only employ image adaptation with complex constraints. In contrast, we alleviate the domain shift from both the image and feature perspectives and validate the effectiveness of each component. Second, they are specifically designed for metric-based predictors, while our SDACD is compatible with most existing deep learning-based methods.

*2.2 Domain adaptation*

Domain adaptation intends to address performance degradation caused by domain shift by transferring knowledge between domains, which benefits the generalization of models. Given the source domain and target domain, most of the existing methods consider unsupervised or semi-supervised situations, where no or only limited labels in the target domain are available. This topic has aroused great interest in recent years and many influential approaches have been proposed [34]-[36]. To the best of our knowledge, domain adaptation mainly focuses on two aspects, namely, image adaptation and feature adaptation.

**Image adaptation** attempts to minimize appearance differences for images of different modals or styles. The intuitive idea is to transform images from one style to another through generation models. The famous CycleGAN [29] first employed bi-directional generation with cycle-consistency constraints in GAN and realized desirable unpaired image-to-image transformation. This provided a reference paradigm for image adaptation. Afterward, many works applied the ideas of CycleGAN to deal with domain shift via image-to-image transformation. For example, CyCADA [30] first employed a cycle-consistent GAN to transform source domain images to the target domain and then trained the model with the synthetic data in the target domain. In contrast, Song et al. [31] transformed target domain images into the source domain, then generated pseudo labels using the source domain model, and finally utilized them to fine-tune the source domain model.

**Feature adaptation** aligns the feature distributions between the source and target domains by exploiting domain-invariant features through adversarial training. DANN [32] was the earliest general framework that employed adversarial training to learn domain-invariant features for domain adaptation. To further improve the discrimination of domain-invariance features, Tzeng et al. [33] combined discriminative models with an untied weight sharing strategy. Recently, Li et al. [34] proposed a novel loss function called maximum density divergence (MDD), which can be combined with adversarial learning to minimize the domain distribution gaps and align the source and target domain on the feature space. Afterward, they also proposed a faster domain adaptation [35] protocol, especially for scenarios with limited training data and energy-sensitive platforms, which executed different number of network layers for different samples according to the adaptation difficulty.

As mentioned above, most existing domain adaptation methods are designed for unsupervised or semi-supervised scenes. We, on the other hand, treat cross-domain change detection as a supervised domain adaptation problem, where there are supervised annotations for bi-temporal images with severe domain shift. Although current methods provide valuable solutions to bridge domain gaps, they suffer from underutilization or loss of image information in cross-domain change detection. Our work is closely related to SIFA [37], which also combines image adaptation and feature adaptation for domain adaptation. SIFA solves the performance degradation by synergistic learning between CT-to-MR transformation and CT segmentation, which mainly focuses on the segmentation accuracy in the CT domain. However, given the pre-event and post-event images share the annotation in change detection, we highlight our supervised domain adaptation framework explores information from all available domains to improve the robustness and precision of the model. This is because change detection model distinguishes changes through feature differences, but these features would suffer more or less information loss during image transformation. In addition, unlike SIFA, we separate image adaptation and feature adaptation as two separate modules, which enables our framework to easily plugged in any non-domain adaptation change detection model. The main contribution of our work is providing a general and flexible change detection framework, helping existing methods handle cross-domain change detection with few adjustments.

## 3  Methodology

In this section, we first introduce the architecture of our proposed framework and then illustrate details of two main modules: the image adaptation module and the feature adaptation module. Finally, we define the full loss function of our framework.

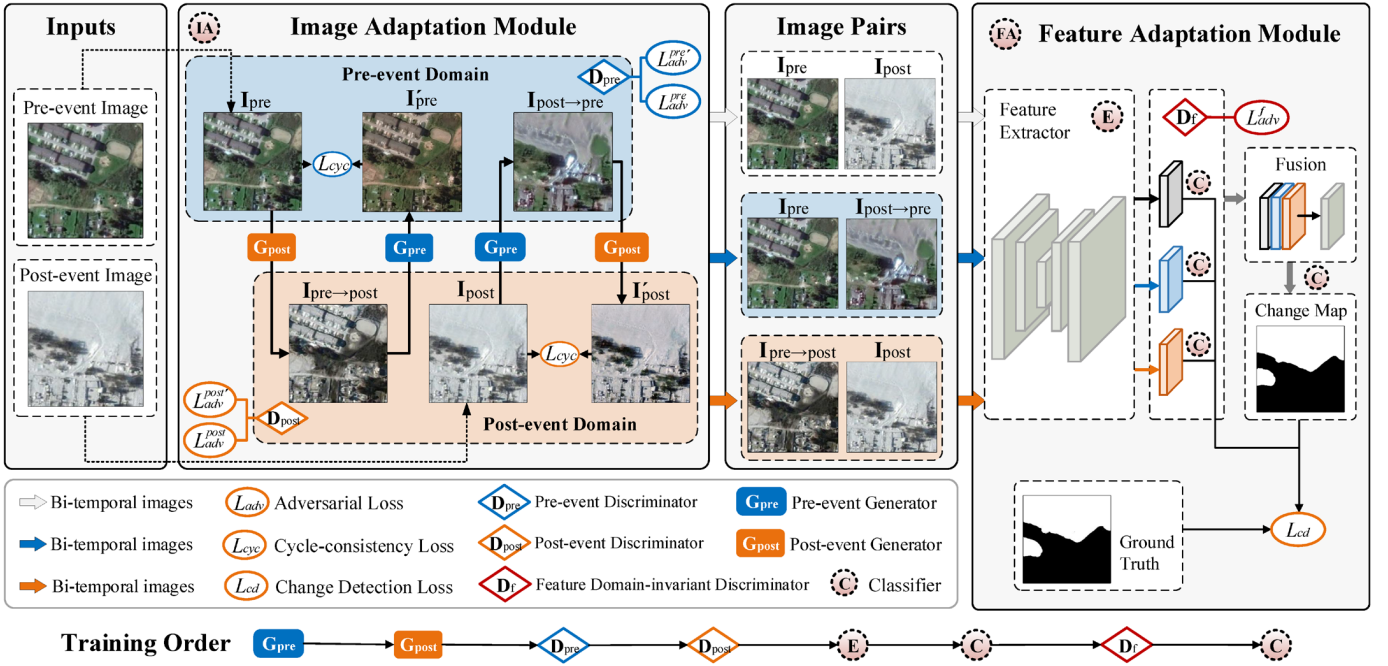

**Fig. 2** The overall architecture of our SDACD framework for cross-domain change detection. Given bi-temporal images ($I_{pre}, I_{post}$) with severe domain shift, we first utilize an image adaptation module (IA) to alleviate the domain shift by aligning the image appearance. Then, we input two bi-temporal images from the pre-event domain and post-event domain, *i.e.*, ($I_{pre}, I_{post \to pre}$) and ($I_{pre \to post}, I_{post}$), as well as the original bi-temporal images ($I_{pre}, I_{post}$) into the feature adaptation module (FA) and extract domain-invariant features via a feature domain-invariant discriminator. The feature adaptation module can further reduce the domain gap by aligning feature distributions from different domains in the feature space. Finally, we integrate the extracted features from three bi-temporal images to predict the final change map by the classifier *C*. We show the training order of key modules in the SDADC framework at the bottom. Better viewed in color.

### 3.1 Overall Architecture

Fig. 2 depicts the overall architecture of our proposed framework. Given the pre-event image $I_{pre} \in R^{H \times W \times C}$ and the post-event image $I_{post} \in R^{H \times W \times C}$, we aim to predict a binary change map $O \in R^{H \times W}$, $\{o_i \in O | o_i \in \{0, 1\}, i = 1, 2, \cdots, HW\}$, where 1 represents changed pixels and 0 means no change. To highlight the cross-domain change detection problem, we denote the pre-event image as $I_{pre} \in PrD$ and the post-event image as $I_{post} \in PoD$ ($PrD \neq PoD$), where $PrD$ and $PoD$ represent the pre-event domain and post-event domain, respectively.

To narrow the domain gap, we design an image adaptation (IA) module and a feature adaptation (FA) module to implement domain adaptation from different perspectives. Accordingly, the pipeline can be divided into two main phases. First, we transform the pre-event image $I_{pre}$ to the post-event domain $PoD$ to obtain the pre-event stylized as post-event image $I_{pre \rightarrow post}$. Similarly, we generate the post-event stylized as pre-event image $I_{post \rightarrow pre}$ by transforming the post-event image $I_{post}$ to the pre-event domain $PrD$. Therefore, we obtain two bi-temporal images in the same domain, where $(I_{pre}, I_{post \rightarrow pre}) \in PrD$, $(I_{pre \rightarrow post}, I_{post}) \in PoD$. Afterward, we input the two generated bi-temporal images $(I_{pre}, I_{post \rightarrow pre})$, $(I_{pre \rightarrow post}, I_{post})$ with the original $(I_{pre}, I_{post})$ into the feature adaptation module, where we further reduce the domain gap by aligning feature distributions in the feature space via a feature domain-invariant discriminator. Finally, to fully utilize the image information from different domains, we predict the final change map $O$ by fusing the aligned features extracted from the above three bi-temporal images. More details of the IA and FA modules will be presented in the following sections.

### 3.2 Image Adaptation Module for Appearance Alignment

Since bi-temporal images are usually captured under different imaging conditions, we specifically design an image adaptation (IA) module to minimize the appearance variations between bi-temporal images, while the original contents with geographical semantics remain unaffected. The goal of the IA module is to learn mapping functions between two domains $PrD$ and $PoD$ given training samples $(I_{pre}, I_{post})$.

As shown in Fig. 2 (IA), we employ generative adversarial networks, which have made wide success in image-to-image transformation, by building two generators $G_{pre}: I_{post} \rightarrow I_{pre}$ and $G_{post}: I_{pre} \rightarrow I_{post}$ and two adversarial discriminators $D_{pre}$ and $D_{post}$. The generator $G_{pre}$ aims to transform images in the post-event domain to the pre-event domain, while $G_{post}$ performs the reverse image transformation. The discriminators $D_{pre}$ and $D_{post}$ compete with their corresponding generators to correctly distinguish the real images from the transformed images. Therefore, the generator $G_{pre}$ and its discriminator $D_{pre}$ can be optimized via adversarial learning:

$$\mathcal{L}_{adv}^{pre}(G_{pre}, D_{\text{pre}}, PoD, PrD) = \mathbb{E}_{I_{pre} \sim PrD}[log D_{pre}(I_{pre})] +$$
$$\mathbb{E}_{I_{post} \sim PoD}[\log(1 - D_{pre}(G_{pre}(I_{post})))] \quad (1)$$

where $G_{pre}$ aims to generate images $G_{pre}(I_{post})$ that have a similar appearance to images from the pre-event domain $PrD$, while $D_{\text{pre}}$ tries to distinguish real images $I_{pre}$ from transformed images $G_{pre}(I_{post})$. $G_{pre}$ aims to minimize this objective against $D_{\text{pre}}$ that tries to maximize it, i.e., $min_{G_{pre}} max_{D_{pre}} \mathcal{L}_{adv}^{pre}(G_{pre}, D_{\text{pre}}, PoD, PrD)$. Similarly, the adversarial loss for the generator $G_{post}$ and its discriminator $D_{post}$ can be defined as:

$$\mathcal{L}_{adv}^{post}(G_{post}, D_{\text{post}}, PrD, PoD) = \mathbb{E}_{I_{post} \sim PoD}[log D_{post}(I_{post})] +$$
$$\mathbb{E}_{I_{pre} \sim PrD}[\log(1 - D_{post}(G_{post}(I_{pre})))] \quad (2)$$

Similar to CycleGAN [29], we employ a reverse generator to impose pixel-wise cycle-consistency to preserve the original geographical elements in the transformed images. As shown in Fig. 2 (IA), for each image $I_{pre}$ from domain $PrD$, the image transformation cycle should be able to bring $I_{pre}$ back to the original image, i.e., $I_{pre} \rightarrow G_{\text{post}}(I_{pre}) = I_{pre \rightarrow post} \rightarrow G_{\text{pre}}(G_{\text{post}}(I_{pre})) = I_{pre \rightarrow post \rightarrow pre} \approx I_{pre}$. Similarly, for each image $I_{post}$ from $PoD$, $G_{\text{pre}}$ and $G_{\text{post}}$ should also satisfy cycle-consistency: $I_{post} \rightarrow G_{\text{pre}}(I_{post}) = I_{post \rightarrow pre} \rightarrow G_{\text{post}}(G_{\text{pre}}(I_{post})) = I_{post \rightarrow pre \rightarrow post} \approx I_{post}$. The cycle-consistency loss $\mathcal{L}_{cyc}$ can be computed as follows:

$$\mathcal{L}_{cyc}(G_{pre}, G_{post}) = \mathbb{E}_{I_{pre} \sim PrD}[||G_{pre}(G_{post}(I_{pre})) - I_{pre}||_1] +$$
$$\mathbb{E}_{I_{post} \sim PoD}[||G_{post}(G_{pre}(I_{post})) - I_{post}||_1] \quad (3)$$

Since our image transformation procedures are cyclical, we also introduce two reverse adversarial losses $\mathcal{L}_{adv}^{pre\prime}$ and $\mathcal{L}_{adv}^{post\prime}$ for the two reverse image transformations, i.e., $I_{pre \rightarrow post} \rightarrow I_{pre \rightarrow post \rightarrow pre} \approx I_{pre}$ and $I_{post \to pre} \to I_{post \to pre \to post} \approx I_{post}$. The reverse adversarial losses $\mathcal{L}_{adv}^{pre'}$ and $\mathcal{L}_{adv}^{post'}$ can be computed as follows:

$$\mathcal{L}_{adv}^{pre'}(G_{pre}, D_{\text{pre}}, PoD, PrD) = \mathbb{E}_{I_{pre} \sim PrD}[\log D_{pre}(I_{pre})] +$$

$$\mathbb{E}_{I_{pre \to post} \sim PoD}[\log(1 - D_{pre}(G_{pre}(I_{pre \to post})))] \quad (4)$$

$$\mathcal{L}_{adv}^{post'}(G_{post}, D_{post}, PrD, PoD) = \mathbb{E}_{I_{post} \sim PoD}[\log D_{post}(I_{post})] +$$

$$\mathbb{E}_{I_{post \to pre} \sim PrD}[\log(1 - D_{post}(G_{post}(I_{post \to pre})))] \quad (5)$$

Therefore, the full adversarial loss of image adaptation $\mathcal{L}_{adv}^{i}$ can be formulated as:

$$\mathcal{L}_{adv}^{i}(G_{pre}, D_{\text{pre}}, G_{post}, D_{\text{post}}, PrD, PoD) = \mathcal{L}_{adv}^{pre}(G_{pre}, D_{\text{pre}}, PoD, PrD) +$$

$$\mathcal{L}_{adv}^{post}(G_{post}, D_{\text{post}}, PrD, PoD) +$$

$$\mathcal{L}_{adv}^{pre'}(G_{pre}, D_{\text{pre}}, PoD, PrD) +$$

$$\mathcal{L}_{adv}^{post'}(G_{post}, D_{\text{post}}, PrD, PoD) \quad (6)$$

*3.3 Feature Adaptation Module for Feature Alignment*

After the original bi-temporal images ($I_{pre}, I_{post}$) are transformed by the above image adaptation module, we obtain several bi-temporal images. Ideally, we can train a change detection model with excellent performance by using bi-temporal images ($I_{pre}, I_{post \to pre}$) or ($I_{pre \to post}, I_{post}$) from the domain $PrD$ or $PoD$. However, for training a cross-domain change detection model with optimal performance, several challenges remain to be solved. First, for bi-temporal images of cross-domain remote sensing, due to the severe domain shift between bi-temporal images, only utilizing image adaptation would be insufficient to achieve the expected domain adaptation results. Second, during the image adaptation procedure, compared with the original image, the generated image has the issue of information loss. Therefore, using only one domain of bi-temporal images to train a change detection model would suffer from underutilization or even loss of image information, thus producing suboptimal results.

Thus motivated, we design a feature adaptation (FA) module to further reduce the remaining domain gap and fully utilize the image information from different domains. Specifically, we impose an additional feature domain-invariance discriminator to align the feature distributions in the feature space and contribute from the perspective of feature adaptation. Generally, feature adaptation aims to extract domain-invariant features without considering the appearance difference between input domains. The most intuitive way is to use a discriminator to distinguish which features come from which domain via adversarial learning. However, since the feature space is high-dimensional, it is difficult to directly align using adversarial learning. Inspired by output space adaptation [38], we enhance the domain invariance of feature distribution via adversarial learning in different lower-dimensional spaces. Specifically, we add adversarial losses between the change detection prediction space and the input bi-temporal image spaces.

As shown in Fig. 2 (FA), we adopt three bi-temporal images from different domains, including $(I_{pre}, I_{post})$ $\in PrD$, $(I_{pre}, I_{post \rightarrow pre}) \in PoD$ and the initial bi-temporal images $(I_{pre \rightarrow post}, I_{post})$, to train the change detection model. We make change predictions for each bi-temporal image and construct a feature domain-invariant discriminator $D_f$ to determine whether the predictions come from $(I_{pre}, I_{post})$, $(I_{pre}, I_{post \rightarrow pre})$, or $(I_{pre \rightarrow post}, I_{post})$. If the features extracted from different domains are aligned, discriminator $D_f$ would fail to differentiate between their corresponding predictions. Otherwise, the adversarial gradients are backpropagated into the feature extractor $E$ of bi-temporal images to minimize the distance among the feature distributions from different domains. Since there are three input bi-temporal images, there are three adversarial losses for feature adaptation, which can be computed as:

$$\mathcal{L}_{adv}^f(I_{pre}, I_{post}, C, E, D_f) = \mathbb{E}[\log D_f(C(E(I_{pre}), E(I_{post})))] +$$
$$\mathbb{E}[\log (1 - D_f(C(E(I_{pre}), E(I_{post}))))] \quad (7)$$

$$\mathcal{L}_{adv}^f(I_{pre \rightarrow post}, I_{post}, C, E, D_f) = \mathbb{E}[\log D_f(C(E(I_{pre \rightarrow post}), E(I_{post})))] +$$
$$\mathbb{E}[\log (1 - D_f(C(E(I_{pre \rightarrow post}), E(I_{post}))))] \quad (8)$$

$$\mathcal{L}_{adv}^f(I_{pre}, I_{post \rightarrow pre}, C, E, D_f) = \mathbb{E}[\log D_f(C(E(I_{pre}), E(I_{post \rightarrow pre})))] +$$
$$\mathbb{E}[\log (1 - D_f(C(E(I_{pre}), E(I_{post \rightarrow pre}))))] \quad (9)$$

where $C$ represents the classifier for change prediction.

Combined with the above three adversarial losses, the full adversarial loss of feature adaptation $\mathcal{L}_{adv}^{f}$ can be formulated as:

$$\mathcal{L}_{adv}^{f}(C, E, D_f) = \mathcal{L}_{adv}^{f}(I_{pre}, I_{post}, C, E, D_f) +$$
$$\mathcal{L}_{adv}^{f}(I_{pre \to post}, I_{post}, C, E, D_f) +$$
$$\mathcal{L}_{adv}^{f}(I_{pre}, I_{post \to pre}, C, E, D_f) \quad (10)$$

As mentioned above, the generated images from image adaptation suffer more or less information loss, which results in a suboptimal model trained with bi-temporal image samples from only one domain. Therefore, for fully utilizing the image information from different domains and keeping training samples diverse, we integrate the aligned features, which are extracted from $(I_{pre}, I_{post})$, $(I_{pre}, I_{post \to pre})$, and $(I_{pre \to post}, I_{post})$, to predict the final change map $O$. We also later conduct ablation studies to investigate the selection of bi-temporal images for training the change detection model and fusion strategies for the prediction final change map, which will be discussed in Section 4.4.2 and Section 4.4.3, respectively.

During the feature adaptation procedure, we make change predictions for each input bi-temporal image and then fuse the aligned features to predict the final change map. Therefore, the full change detection loss $\mathcal{L}_{CD}$ can be computed as follows:

$$\mathcal{L}(O, Y) = \mathcal{L}_{wce}(O, Y) + \mathcal{L}_{dice}(O, Y) \quad (11)$$

$$\mathcal{L}_{CD} = \sum_{i=0,1,2} \mathcal{L}(\hat{O}_i, Y) + \mathcal{L}(O, Y) \quad (12)$$

where $i$ is the index of the input bi-temporal images, $\hat{O}_i$ is the predicted result of the $i$-th bi-temporal images for the feature adaptation, $O$ is the final prediction of our framework, and $Y$ is the ground truth. $L(\cdot)$ is a hybrid loss function for SNUNet [17], which consists of the weighted cross-entropy loss $\mathcal{L}_{wce}(\cdot)$ and the dice loss $\mathcal{L}_{dice}(\cdot)$. Note that any loss function can be used in our framework, and the loss function *(11)* is only employed in SNUNet-SDACD. For a fair comparison, we set the same loss function as their corresponding baselines in our experiments.

*3.4 Full Objective*

The key characteristic of SDACD is that both image and feature adaptations are unified into an end-to-end trainable framework. In each training iteration, the key modules are sequentially updated in the following order: $G_{pre} \rightarrow G_{post} \rightarrow D_{pre} \rightarrow D_{post} \rightarrow E \rightarrow C \rightarrow D_f \rightarrow C$. Specifically, in the image adaptation module, the generators $G_{pre}$ and $G_{post}$ are updated first to transform $I_{post}$ and $I_{pre}$ into generated pre-event and post-event images, *i.e.*, $I_{post \rightarrow pre}$ and $I_{pre \rightarrow post}$, respectively. Then, the discriminators $D_{pre}$ and $D_{post}$ are updated to distinguish the real images from the generated images. Next, the feature extractor $E$ in the feature adaptation module is updated to extract features from the three input bi-temporal images, and then the change predictor $C$ is updated to predict change maps for each bi-temporal image based on the extracted features. Afterward, the feature domain-invariance discriminator $D_f$ is updated to distinguish the input domain of the three prediction maps. Finally, the change predictor $C$ is updated to predict the final change map based on the fused features. Therefore, combined with the cycle-consistency loss and adversarial loss of image adaptation in Eq. (3) and Eq. (6), the full adversarial loss of feature adaptation in Eq. (10) and the change detection loss in Eq. (12), the full multitask loss function can be defined as:

$$\mathcal{L} = \lambda_{cyc}\mathcal{L}_{cyc} + \lambda_i \mathcal{L}_{adv}^i + \lambda_f \mathcal{L}_{adv}^f + \lambda_{CD}\mathcal{L}_{CD} \tag{13}$$

Where we conducted the grid search for the hyperparameters and empirically set $\lambda_{cyc}$=10, $\lambda_i = 1$, $\lambda_f = 0.1$, and $\lambda_{CD} = 1$ in all experiments.

Furthermore, another characteristic of our SDACD is its universality, as we can integrate most non-domain adaptation networks into our framework with few modifications. Given a non-domain adaptation change detection network consisting of a feature extractor $E'$ and a classifier $C'$, we can directly replace $E$ and $C$ in our pipeline, as shown in Fig. 2, which enables the original network to tackle the domain shift between bi-temporal images.

## 4  Experiments

In this section, we implement the SDACD framework based on STANet (2020), DASNet (2021), and SNUNet (2021) and conduct experiments on the CDD and WHU building datasets, which contain time-varying changes caused by seasons and luminance. We compare our method with the baseline as well as state-of-the-art methods and carefully analyze the functions of each proposed module.

## 4.1 Datasets and Evaluation Metrics

The **CDD** [21] is a widely used dataset for change detection proposed by M. A et al., and it contains 11 pairs of bi-temporal images obtained from Google Earth in different seasons with a spatial resolution ranging from 3 to 100 cm per pixel. Since the original bi-temporal images included seven 4725×2700 image pairs and four 1900×1000 image pairs, which were too large to process directly, [21] cropped the original image pairs into the same size of 256×256 pixels, thus generating 10000 image pairs for training, 3000 image pairs for validation, and 3000 image pairs for testing. As shown in Fig. 1, the appearances of bi-temporal images vary greatly due to different seasons. To conform to the assumption $I_{pre} \in PrD, \; I_{post} \in PoD$ where $PrD \neq PoD$, we partially adjusted the order of bi-temporal image pairs to obtain similar appearances in each domain, forming summer-styled pre-event images and winter-styled post-event images. The annotations of changed regions consist of objects varied in size and category, including cars, roads, constructions, etc.

The **WHU Building Dataset** [22] focuses on the changes in buildings and contains one pair of very high-resolution images sized 32507×15354 pixels with a spatial resolution of 0.075 m per pixel. [22] divided the original image pair into a training set of 21243×15354 pixels and a testing set of 11265×15354 pixels. Because of memory constraints, we further cropped the images into patches of 256×256 pixels without overlap thus producing 4980 bi-temporal images for training and 2700 bi-temporal images for testing. Note that images were taken in 2012 and 2016 separately. Thus, bi-temporal images showed varied appearances caused by luminance fluctuations, as shown in Fig. 1. The annotations only consist of changes caused by the construction and destruction of buildings.

We adopt three metrics to evaluate the performance of cross-domain change detection, including Precision (*P*), Recall (*R*), and *F1-score*. While Precision reflects the accuracy of detection, Recall represents the completeness of the predicted changed regions. Generally, the *F1-score* is of more concern to researchers because it takes both precision and recall into account and reflects the comprehensive performance of the method.

### 4.2 Implementation details

Our experiments were conducted on a high-performance computing server with NVIDIA Tesla V100 (16G) GPUs. All models were trained with 4 GPUs and evaluated with 1 GPU. We implemented our SDACD framework based on three baseline models, denoted as STANet-SDACD, DASNet-SDACD, and SNUNet-SDACD. Following their original settings in [17][18][19], we initialized STANet and DASNet with publicly available pretrained ImageNet parameters, while SNUNet was initialized by kaiming normalization. The other parameters were all initialized by kaiming normalization. For data augmentation, we employed random flips, random rotation, and normalization. In addition, since the proportion of changed areas is relatively small in the WHU building dataset, we further augmented the training set with 3442 bi-temporal images generated by randomly cropping several 256×256 patches around each instance of changed areas. We adopted the Adam optimizer, and the initial learning rates were set to 5e-4 in all experiments. We trained STANet and STANet-SDACD for 200 epochs, DASNet and DASNet-SDACD for 60 epochs, and SNUNet and SNUNet-SDACD for 120 epochs. Note that all networks were trained in an end-to-end manner.

### 4.3 Comparison with State-of-the-Art Methods

#### 4.3.1 Comparison Methods

To demonstrate the superiority of our framework, we compared the performance of our framework with several representative and state-of-the-art deep learning-based change detection methods.

**FC-EF, FC-Siam-conc** and **FC-Siam-diff** [16] are the earliest FCN-based methods, which integrated UNet into the Siamese architecture for change detection. These three methods explored three different fusion strategies, including bi-temporal image fusion, multilevel feature concatenation, and multilevel feature difference, which are the main strategies for bi-temporal feature fusion in existing change detection methods.

**STANet** [18] is an impressive work that utilizes a pyramid transformer structure to learn multilevel spatial-temporal dependencies. This structure effectively eliminates incorrect detections caused by luminance differences and misregistration.

**DASNet** [19] employs a Siamese convolutional module that obtains local features of the images at different times and then builds connections between local features by a dual attention module, which exploits global contextual information to distinguish the changed areas from the unchanged ones.

**SNUNet** [17] is the current state-of-the-art method on the CDD dataset. It employs a UNet++ architecture to extract multilevel representations and an elaborate ensemble channel attention module (ECAM) to integrate multilevel features, which boosts the performance to a higher level.

*4.3.2 Performance on CDD Dataset*

The second column of Table 1 list the experimental results on the CDD dataset. Although STANet, DASNet, and SNUNet have already exhibited impressive performance, our SDACD-STANet, SDACD-DASNet, and SDACD-SNUNet can bring consistent improvements on the *F1-score* by 0.7%, 0.78%, and 1.66%, respectively. Notably, based on the cutting-edge work SNUNet, our method achieves new state-of-the-art performance on CDD with an *F1-score* of 97.34%. Our framework outperforms the baseline model and mainly benefits from the consideration of domain adaptation from two perspectives. First, the IA module aligns the image appearance by image-to-image transformation, which effectively relieves the effect of different seasons and climates on the land cover changes. Second, the FA module further reduces the remaining domain shift by exploiting domain-invariance features to predict the final change map. In addition, the results in Table 1 also imply that the performance of other deep learning-based models could be further enhanced by combining them with our SDACD framework. Therefore, our framework shows great potential in cross-domain change detection as an easy-to-plug-in module, which provides a simple but effective way to tackle the performance degradation caused by different seasons and climates for existing non-domain adaptation models.

As shown in Fig. 3, since bi-temporal images have significantly different appearances (summer *vs.* winter), this poses great challenges for the baseline models, which do not consider domain shift, to discriminate real changes from false-positives, especially for changed regions with delicate structures. However, our SDACD can minimize the impact of time-varying land cover changes and accurately detect fine boundaries of roads and small-scale object changes. This explicitly demonstrates the superiority of our framework for cross-domain change detection.

**Table 1.** Performance comparison with the state-of-the-art methods on the CDD and WHU building datasets. The red and blue arrows indicate the increase and decrease in metrics compared with the baseline model, respectively.

| Methods | CDD | | | WHU building | | |
|---|---|---|---|---|---|---|
| | P (%) | R (%) | F (%) | P (%) | R (%) | F (%) |
| FC-EF | 84.68 | 65.13 | 73.63 | 80.75 | 67.29 | 73.40 |
| FC-Siam-diff | 87.57 | 66.69 | 75.07 | 48.84 | 88.96 | 63.06 |
| FC-Siam-conc | 88.81 | 62.20 | 73.16 | 54.20 | 81.34 | 65.05 |
| STANet | 83.17 | 92.76 | 87.70 | 82.12 | 89.19 | 83.40 |
| SDACD-STANet | 87.40 ↑4.23 | 89.50 ↓3.26 | 88.40 ↑0.70 | 90.90 ↑8.78 | **93.50** ↑4.31 | 92.21 ↑8.81 |
| DASNet | 93.28 | 89.91 | 91.57 | 83.77 | 91.02 | 87.24 |
| SDACD-DASNet | 92.85 ↓0.43 | 91.87 ↑1.96 | 92.35 ↑0.78 | 89.21 ↑5.44 | 90.46 ↓0.56 | 89.83 ↑2.59 |
| SNUNet | 96.60 | 94.77 | 95.68 | 82.12 | 89.19 | 85.51 |
| SDACD-SNUNet | **97.13** ↑0.53 | **97.56** ↑2.79 | **97.34** ↑1.66 | **93.85** ↑11.73 | 90.91 ↑1.72 | **92.36** ↑6.85 |

**Table 2.** Comparison of model size and computational complexity with baseline models.

| Models | Params/M | FLOPs/G |
|---|---|---|
| STANet | 16.93 | 26.33 |
| SDACD-STANet | 40.05 | 286.68 |
| DASNet | 48.22 | 201.44 |
| SDACD-DASNet | 80.94 | 752.82 |
| SNUNet | 27.07 | 246.32 |
| SDACD-SNUNet | 52.92 | 1368.48 |

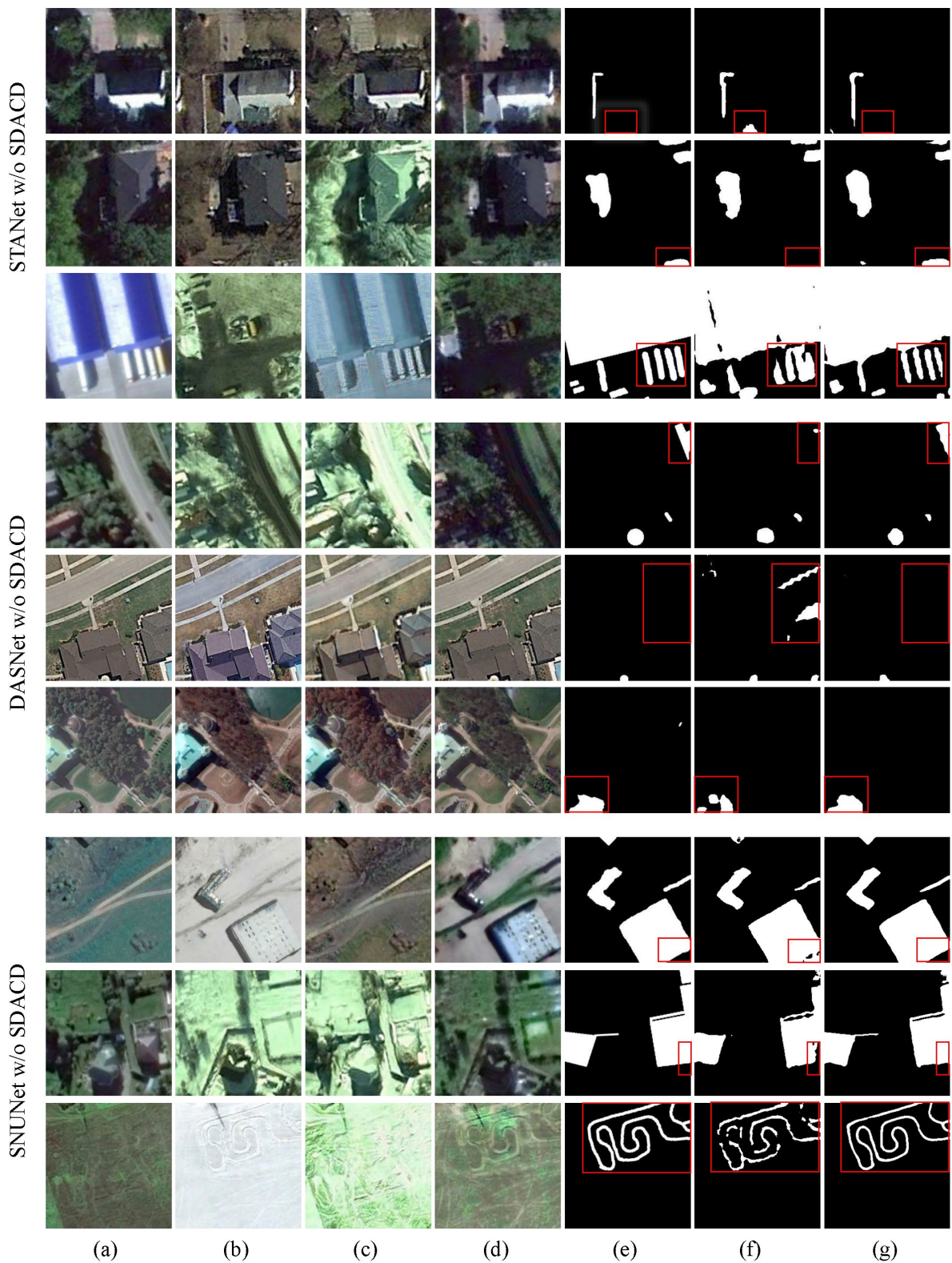

**Fig. 3** Visualized experimental results on CDD. Images from left to right are (a) pre-event images, (b) post-event images, (c) pre-event stylized as post-event images, (d) post-event stylized as pre-event images, (e) ground truth, (f) the results of baseline, and (g) the results of baseline + SDACD. Compared with the corresponding baselines, our method can effectively avoid false-positive predictions introduced by varied seasons and climates and preserve detailed changes with clearer boundaries.

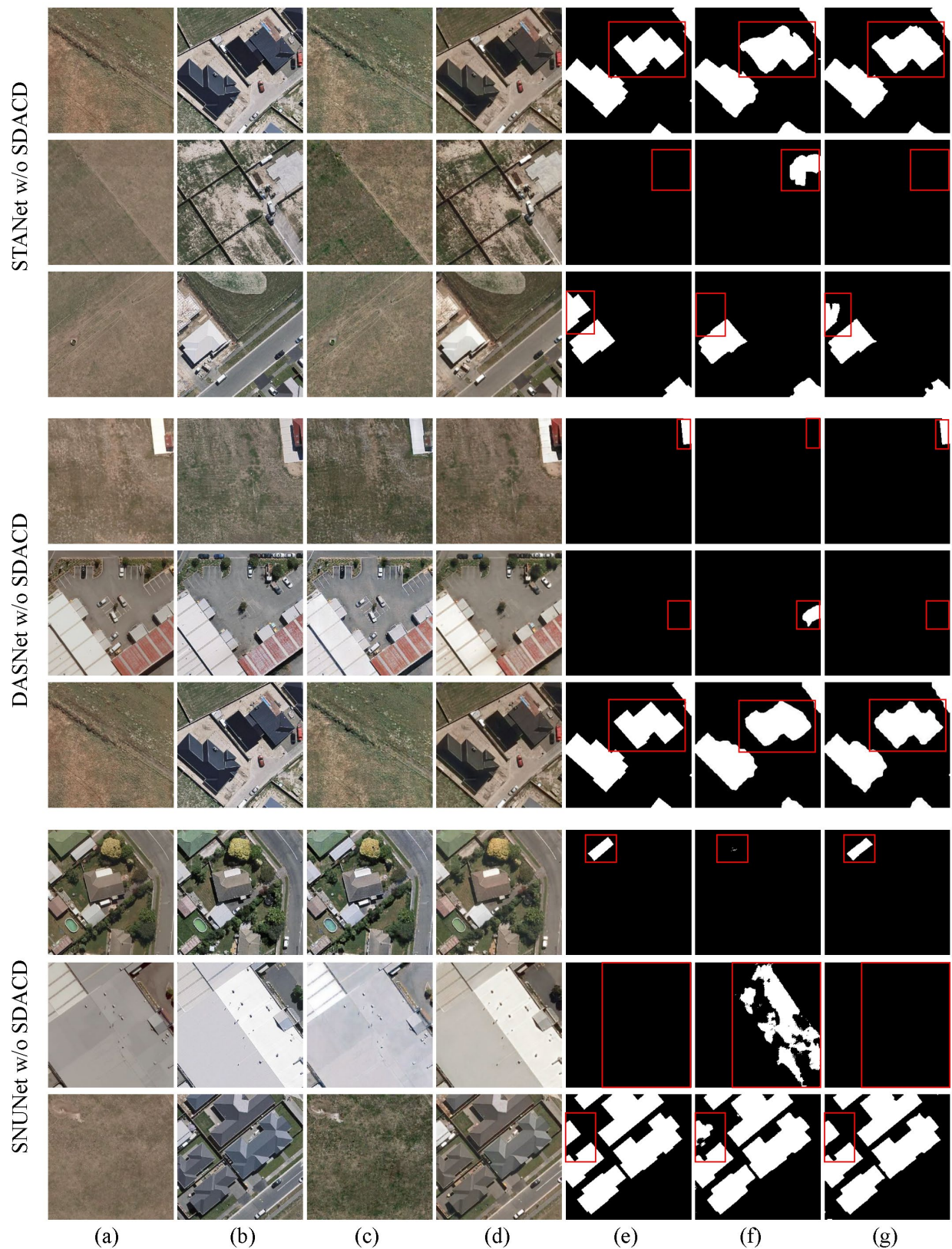

**Fig. 4.** Visualized experimental results on the WHU Building Dataset. Images from left to right are (a) pre-event images, (b) post-event images, (c) pre-event stylized as post-event images, (d) post-event stylized as pre-event images, (e) ground truth, (f) the results of baseline, (g) the results of baseline + SDACD.

*4.3.3 Performance on the WHU Building Dataset*

Different from CDD, which is a class-agnostic change detection dataset, the WHU building dataset focuses on building changes. Therefore, the WHU building dataset is considered to be more challenging than CDD. We further conducted experiments on the WHU building dataset to see if our framework can work under this more challenging scenario. We report the experimental results in the third column of Table 1. Similar to CDD, our framework outperforms the baseline model by a large margin, bringing an increase in the *F1-score* of 8.81% on STANet, 2.59% on DASNet, and 6.85% on SNUNet. Moreover, our SDACD-SNUNet also achieves new state-of-the-art performance on the WHU building dataset with an *F1-score* of 92.36%. These results further confirm that our framework can not only effectively bridge the appearance difference by image adaptation but also provide domain-invariant and discriminative representations, which are significant for the precision of change detection.

Visual comparison results are provided in Fig. 4. We notice that without domain adaptation, the baseline models are sensitive to pseudo changes caused by luminance fluctuations. In contrast, since our framework provides domain adaptation ability to the baseline model, it helps to avoid false-positives and false-negatives, thus generating more precise change predictions with clear boundaries and enhancing the performance of cross-domain change detection.

In addition, from the performance comparison results in Table 1, we note that although our SDACD implementations consistently achieves higher *F1-score*, in some cases, the precision and recall of SDACD-STANet and SDACD-DASNet decline. For the change detection task, we usually focus more on the performance metric of *F1-score*, and this degradations of *precision* or *recall* can be regarded as a trade-off. Because higher *precision* would lead to lower *recall* as the threshold increases. Specifically, the degradations of SDACD-DASNet on both CDD and WHU building datasets are rather small and negligible, and the decrease of *recall* of SDACD-STANet on CDD is mainly caused by false negatives of small-scale changes with higher threshold settings.

*4.3.4 Model Size and Computational Complexity*

We further compare the number of parameters and computational efficiency among three baseline models and our SDACD implementations in Table 2. Since we introduce an IA module and an FA module to alleviate the domain gap, the efficiency is sacrificed for better performance under cross-domain scenarios. Compared with the baseline models, i.e., STANet, DASNet, and SNUNet, the parameters of our SDACD implementations increase by 23.12M, 32.72M, and 25.85M, respectively. The parameter increments are relatively fixed. However, the increment of FLOPs is highly related to the computational complexity of the baseline model. This is because we repeatedly employ the feature extractor of the baseline model to extract feature maps from three pairs of bi-temporal images, and then integrate these features from different domains to learn more discriminative representations, as illustrated in Fig. 2 (FA). Note that the main purpose of this paper is to propose an effective solution for the challenging cross-domain change detection, so we leave designing a more efficient framework as future work.

*4.4 Ablation Study*

In this subsection, we report the results of ablation studies on the WHU building dataset to evaluate the effectiveness of each component proposed in SDACD based on SNUNet. In addition, we also reveal insights into the framework design, *i.e.,* why we employ three bi-temporal images for feature extraction and which strategy is the best for fusing feature maps from different domains.

*4.4.1 Effectiveness of IA and FA Modules*

We first conduct experiments to verify the effectiveness of the proposed IA and FA modules. The experimental results are reported in Table 3 and demonstrate that both the IA module and FA module improve the performance of the baseline model by a large margin. Specifically, with the help of the IA module, our framework significantly improves the precision by 9.73%, which contributes most to the improvement in the *F1-score*. As shown in Fig. 5, this is because our IA module can relieve the appearance difference between bi-temporal images and effectively avoid most false positives produced by the baseline model. Moreover, it can be observed that incorporating the FA module with the IA module achieves the best performance with a competitive recall value and an improvement of 2% in precision. The reason is that, as described in Section 3.3, image adaptation alone is insufficient for the optimal results due to severe domain shift between bi-temporal images and more or less information loss during image-to-image transformation. Thus, the FA module further improves the performance by exploiting domain-invariance features to alleviate the domain gap from another perspective. The visualized results in Fig. 6 also reveal similar observations. Therefore, we can conclude that our proposed adaptation modules, *i.e.*, IA and FA, are both effective in boosting the cross-domain change detection performance.

Table 3. Impact of the proposed IA and FA modules on the WHU building dataset.

| Baseline | IA | FA | P (%) | R (%) | F (%) |
|---|---|---|---|---|---|
| | | | 82.12 | 89.19 | 85.51 |
| SNUNet | ✓ | | 91.85 | **91.10** | 91.47 |
| | ✓ | ✓ | **93.85** | 90.91 | **92.36** |

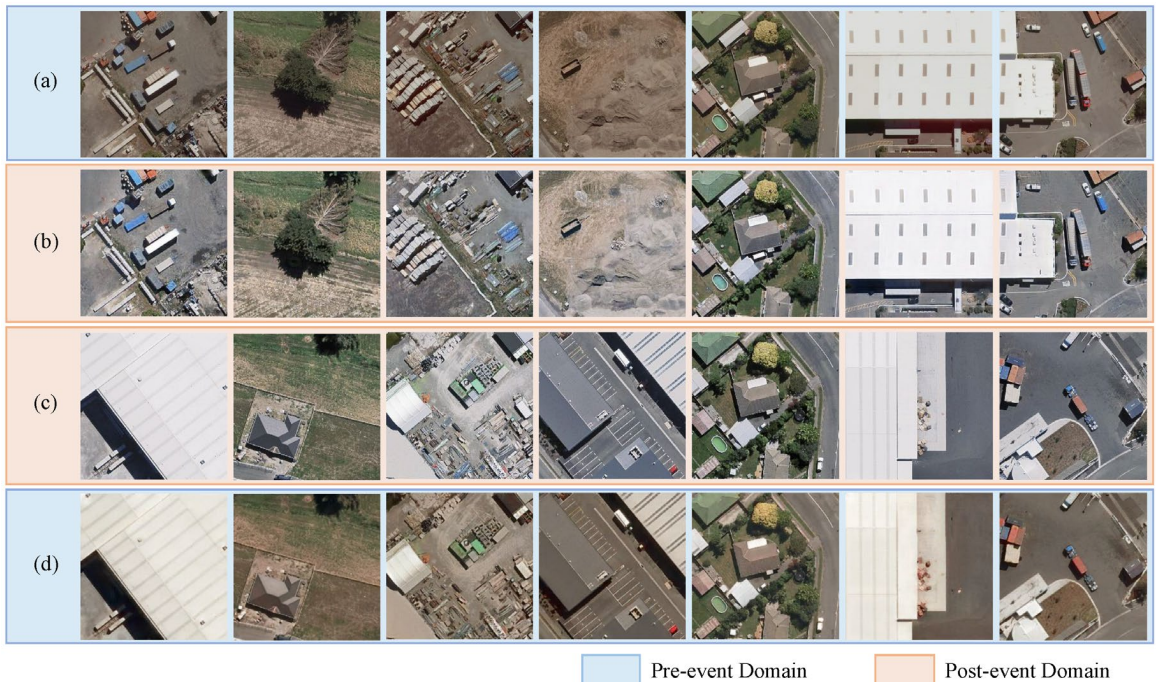

Fig. 5. The results of cross-domain image-to-image transformation through the IA module. Images from top to bottom are (a) pre-event images, (b) pre-event stylized as post-event images, (c) post-event images, and (d) post-event stylized as pre-event images. Note that images of (a) and (d) belong to the pre-event domain, while (b) and (c) belong to the post-event domain. The transformed images in the same domain share a similar appearance, which validates the effectiveness of the IA module.

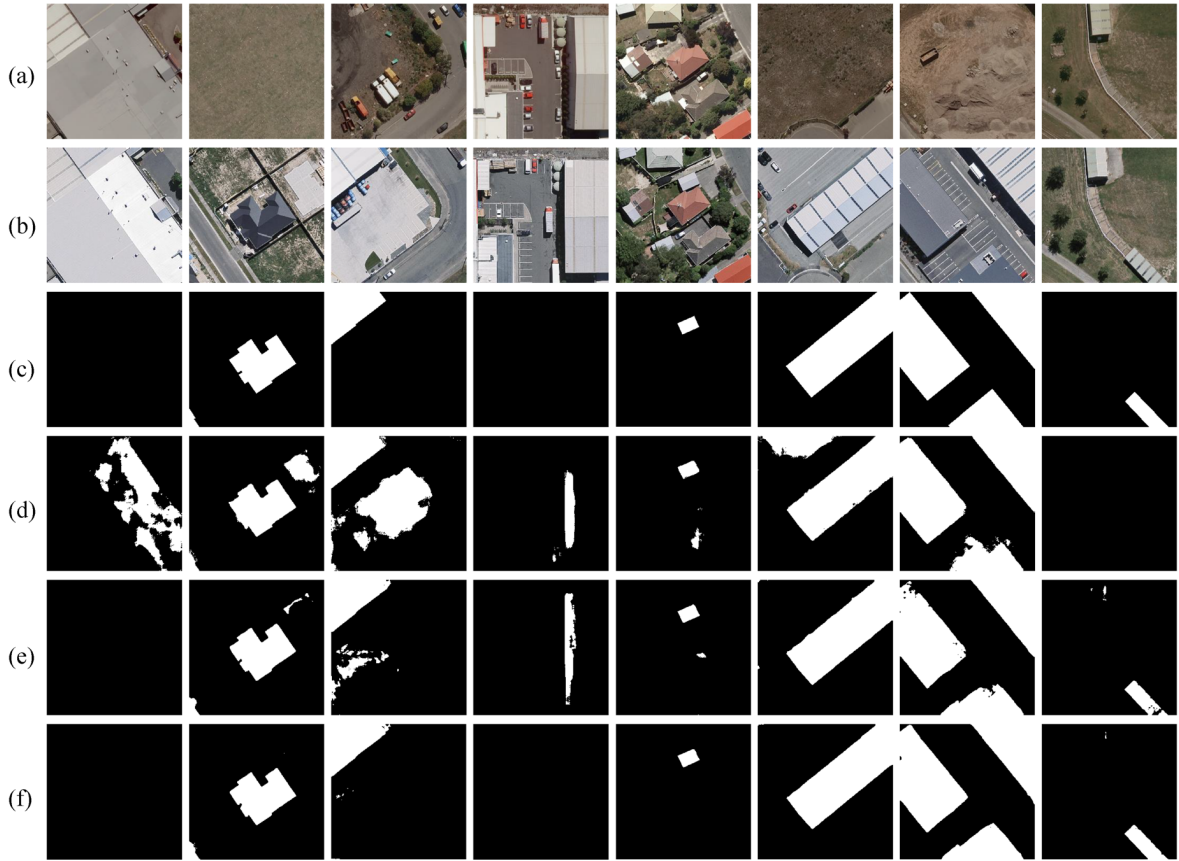

**Fig. 6.** Visualized change maps of the ablation studies for IA and FA modules. Images from left to right are (a) pre-event images, (b) post-event images, (c) ground truth, (d) the results of SNUNet, (e) the results of SNUNet + IA, (f) the results of SNUNet + IA + FA. Both IA and FA help to reduce false-positive predictions and generate fine details for the changed regions. The best performance is achieved when combining the baseline model with IA and FA.

*4.4.2 Impact of Different Combinations of Bi-temporal Images*

Different from general domain adaptation tasks [29][30][31] where only source-domain labels are available, for cross-domain change detection, the transformed bi-temporal images are still considered to share the same annotations of changed regions. This makes it possible to adopt information from different domains for predictions. On the other hand, the generated bi-temporal images suffer from information loss more or less. Thus, using only one domain bi-temporal images to train the change detection model would underutilize the image information. Therefore, we further explore the impact of different combinations of bi-temporal images.

Intuitively, we experiment on three pairs of bi-temporal images, *i.e.,* the original ($I_{pre}, I_{post}$), ($I_{pre}, I_{post \rightarrow pre}$) ∈ $PrD$ and ($I_{pre \rightarrow post}, I_{post}$) ∈ $PoD$, which results in seven different combinations in total.

We present the results in Table 4 and visualize the change maps in Fig. 7 for comparison. As observed, when only one pair of bi-temporal images is used, the performance is $(I_{pre}, I_{post}) < (I_{pre}, I_{post \rightarrow pre}) < (I_{pre \rightarrow post}, I_{post})$, which proves that relieving appearance differences contributes to cross-domain change detection. If we combine two or more pairs of bi-temporal images to extract feature maps, the performance shows consistent improvements. When exploiting all three pairs of bi-temporal images, we obtain the best performance on the WHU building dataset with an $F1$-$score$ of 92.36%, which is 6.85% higher than the baseline. Here, we emphasize that incorporating the original bi-temporal images $(I_{pre}, I_{post})$ can also benefit change detection because they contain all available information. Another interesting observation is that if one pair of bi-temporal images obtains higher scores alone, such bi-temporal images can bring more obvious improvements when combined with other pairs. This provides guidance to select proper combinations of bi-temporal images.

Moreover, although we achieve the best performance with the aforementioned combination of three bi-temporal images, we highlight that not all pairs of bi-temporal images as well as generated images can constantly boost the performance. We also conducted experiments combining $(I_{pre \rightarrow post}, I_{post \rightarrow pre})$ and possible combinations of reconstructed bi-temporal images $(I_{pre \rightarrow post \rightarrow pre}, *)$ and $(I_{post \rightarrow pre \rightarrow post}, *)$, but the performance of the change detection model was worse than that of even the original bi-temporal images. The reason could be that those bi-temporal images suffer from information loss after iterative image-to-image transformation, which is harmful for the final change prediction.

**Table 4.** Comparison of different combinations of bi-temporal images on the WHU building dataset.

| $I_{pre} + I_{post}$ | $I_{pre} + I_{post\text{-}pre}$ | $I_{pre\text{-}post} + I_{post}$ | P (%) | R (%) | F (%) |
|---|---|---|---|---|---|
| ✓ | | | 82.12 | 89.19 | 85.51 |
| | ✓ | | 87.62 | 86.94 | 87.28 |
| | | ✓ | 92.10 | 90.19 | 91.14 |
| ✓ | ✓ | | 87.28 | 88.09 | 87.68 |
| ✓ | | ✓ | 92.35 | 90.76 | 91.55 |
| | ✓ | ✓ | 91.91 | 90.62 | 91.26 |
| ✓ | ✓ | ✓ | **93.85** | **90.91** | **92.36** |

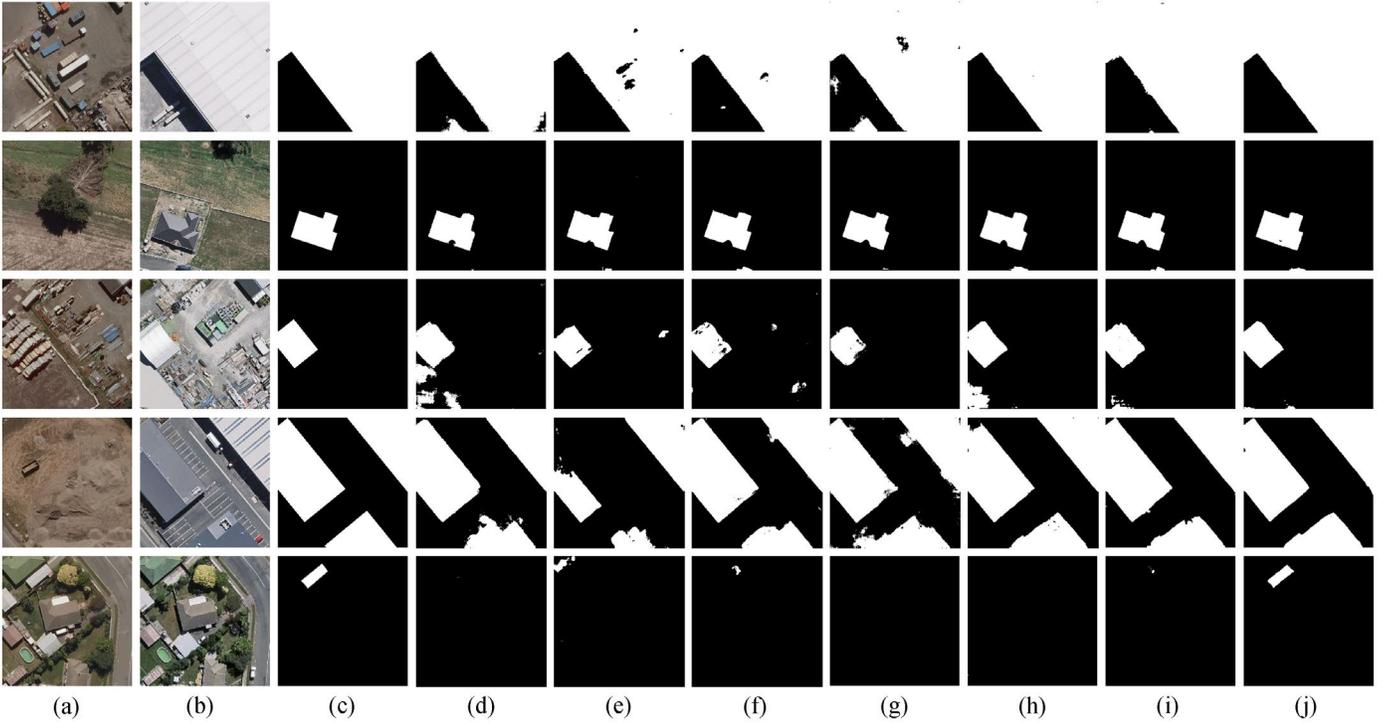

**Fig. 7.** Comparison of visualized change maps produced by different combinations of bi-temporal images. For simplicity, we denote the bi-temporal images of $(I_{pre}, I_{post})$, $(I_{pre}, I_{post \rightarrow pre})$ and $(I_{pre \rightarrow post}, I_{post})$ as I, II and III, respectively. Images from left to right are (a) pre-event images, (b) post-event images, (c) ground truth, the results of (d) I, (e) II, (f) III, (g) I + II, (h) I + III, (i) II + III, and (j) I + II + III, respectively.

### 4.4.3 Discussion of Different Fusion Strategies

Since we obtain three groups of domain-invariant feature maps after feeding three pairs of bi-temporal images into the feature adaptation module, we are curious as to how to integrate these feature maps to produce better results. Here, we explore two different fusion strategies, *i.e.*, output fusion and feature fusion, which are widely used in the computer vision community [39][40]. Output fusion refers to first predicting the change map for each feature of bi-temporal images and then integrating the results, while feature fusion integrates all feature maps for the final prediction. As shown in Table 5, feature fusion yields an *F1-score* of 92.36%, which is 0.5% higher than the output fusion. As shown in Fig. 8, we observe that feature fusion can discriminate detailed changes in buildings and effectively prevent false-positive predictions. We consider that feature fusion can better utilize the image information from different domains and compensate for each other, which formulates more expressive representations. Therefore, we employ feature fusion in our FA module, where three groups of feature maps from different domains are concatenated first and then fused to predict the final change map.

Table 5. Comparison of different fusion strategies on the WHU building dataset.

| Fusion Strategy | P (%) | R (%) | F (%) |
|---|---|---|---|
| output fusion | 93.21 | 90.55 | 91.86 |
| feature fusion | **93.85** | **90.91** | **92.36** |

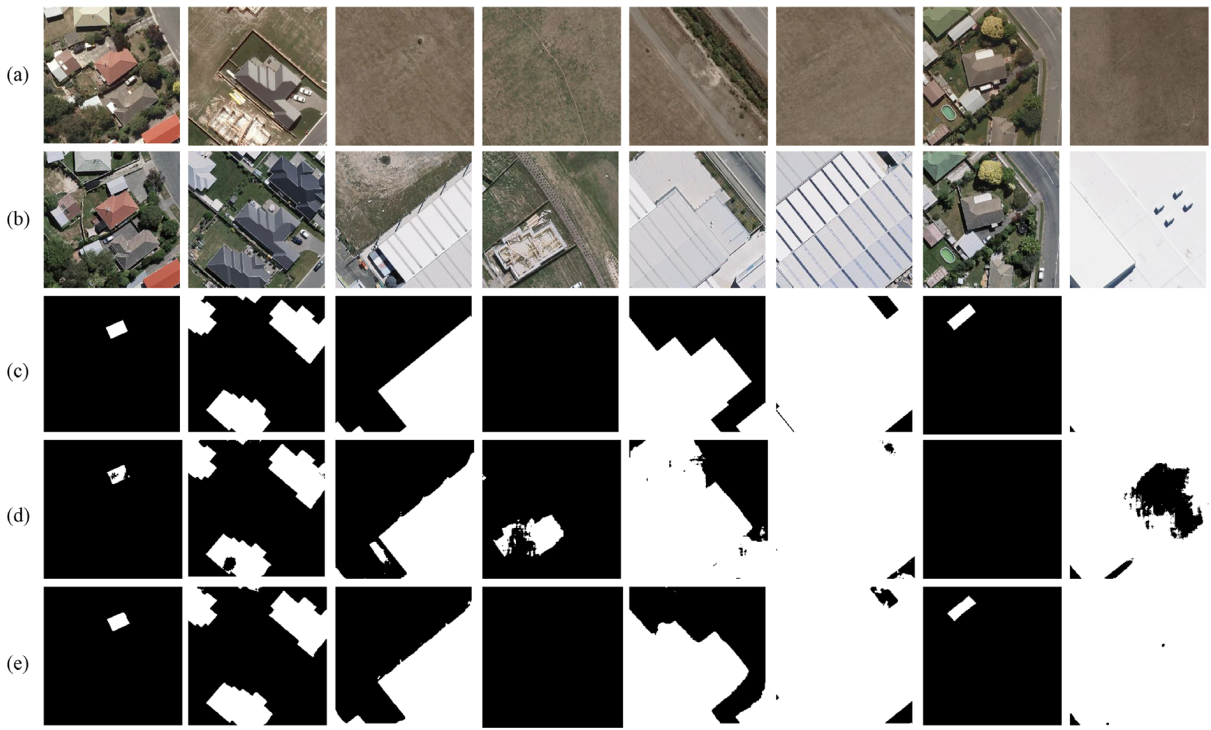

**Fig. 8.** Comparison of the change maps with different fusion strategies. Images from top to bottom are (a) pre-event images, (b) post-event images, (c) ground truth, (d) the results of output fusion, and (e) the results of feature fusion. Feature fusion is shown to be more powerful in capturing semantic information than output fusion, producing complete change regions with accurate boundaries owing to integrated representations.

## 5 Conclusion

In this paper, we propose a novel end-to-end supervised domain adaptation framework called SDACD, which can be easily incorporated into non-domain adaptation networks to tackle challenging cross-domain change detection on very high-resolution remote sensing images. The proposed IA module leverages generative adversarial networks to realize cross-domain image-to-image transformation, which aligns the appearance of bi-temporal images. To further reduce the domain gap, we introduce an FA module, which extracts domain-invariant features from different domains for feature alignment. And we integrate feature maps from all domains for the final change prediction. Our experiments show the superiority of our framework in overcoming domain shift. It achieves new state-of-the-art *F1-scores* of 97.34% and 92.36% on the CDD and WHU building datasets, respectively. More importantly, our method provides a general solution with great potential that is capable of bringing consistent increments to existing change detection methods and friendly to real-world applications for change detection.

Since we add an image adaptation (IA) module and a feature adaptation (FA) module to solve the domain shift of cross-domain change detection, our proposed framework sacrifices some efficiency compared with the baseline. In future work, it would be promising to consider compressing our framework through knowledge distillation to improve efficiency, which can further benefit real-world applications.

**Acknowledgments**

This work was supported in part by the National Natural Science Foundation of China under Grants 62076186 and 61822113 and in part by the Science and Technology Major Project of Hubei Province (Next-Generation AI Technologies) under Grant 2019AEA170. The numerical calculations in this paper have been done performed on the supercomputing system in the Supercomputing Center of Wuhan University.

**References**

[1] Haobo Lyu, Hui Lu, Lichao Mou, Wenyu Li, Jonathon Wright, Xuecao Li, Xinlu Li, Xiao Xiang Zhu, Jie Wang, Le Yu, Peng Gong. 2018. Long-term annual mapping of four cities on different continents by applying a deep information learning method to landsat data. *Remote Sensing* 10, 3 (2018), 471.

[2] Yunhao Gao, Feng Gao, Junyu Dong, and Shengke Wang. 2019. Transferred deep learning for sea ice change detection from synthetic-aperture radar images. *IEEE Geoscience and Remote Sensing Letters* 16, 10 (2019), 1655–1659.

[3] Sudipan Saha, Francesca Bovolo, and Lorenzo Bruzzone. 2018. Destroyed-buildings detection from VHR SAR images using deep features. In *Image and Signal Processing for Remote Sensing XXIV*, Vol. 10789. International Society for Optics and Photonics, 107890Z.

[4] William A Malila. 1980. Change vector analysis: an approach for detecting forest changes with Landsat. *In LARS symposia*. 385.

[5] JS Deng, K Wang, YH Deng, and GJ Qi. 2008. PCA-based land-use change detection and analysis using multitemporal and multisensor satellite data. *International Journal of Remote Sensing* 29, 16 (2008), 4823–4838.

[6] Prashanth R Marpu, Paolo Gamba, and Morton J Canty. 2011. Improving change detection results of IR-MAD by eliminating strong changes. *IEEE Geoscience and Remote Sensing Letters* 8, 4 (2011), 799–803. 5-7

[7] Chen Wu, Bo Du, Xiaohui Cui, and Liangpei Zhang. 2017. A post-classification change detection method based on iterative slow feature analysis and Bayesian soft fusion. *Remote Sensing of Environment* 199 (2017), 241–255.

[8]  Ashish Ghosh, Niladri Shekhar Mishra, and Susmita Ghosh. 2011. Fuzzy clustering algorithms for unsupervised change detection in remote sensing images. *Information Sciences* 181, 4 (2011), 699–715.

[9]  Konrad J Wessels, Frans Van den Bergh, David P Roy, Brian P Salmon, Karen C Steenkamp, Bryan MacAlister, Derick Swanepoel, and Debbie Jewitt. 2016. Rapid land cover map updates using change detection and robust random forest classifiers. *Remote Sensing* 8, 11 (2016), 888.

[10] Hassiba Nemmour and Youcef Chibani. 2006. Multiple support vector machines for land cover change detection: An application for mapping urban extensions. *ISPRS Journal of Photogrammetry and Remote Sensing* 61, 2 (2006), 125–133.

[11] Jun Yu, Dacheng Tao, Meng Wang, and Yong Rui. 2015. Learning to Rank using User Clicks and Visual Features for Image Retrieval. *IEEE Transactions on Cybernetics*, 45(4):767-779.

[12] Chaoqun Hong, Jun Yu, Jian Wan, Dacheng Tao, and Meng Wang. 2015. Multimodal Deep Autoencoder for Human Pose Recovery. *IEEE Transactions on Image Processing,* 24(12):5659-5670.

[13] Chaoqun Hong, Jun Yu, Jian Zhang, Xiongnan Jin, and Kyong-Ho Lee. 2019. Multimodal face-pose estimation with multitask manifold deep learning. *IEEE Transactions on Industrial Informatics*, 15(7): 3952-3961.

[14] Jun Yu, Min Tan, Hongyuan Zhang, Yong Rui, and Dacheng Tao. 2022. Hierarchical deep click feature prediction for fine-grained image recognition. *IEEE Transactions on Pattern Analysis and Machine Intelligence*, 44(2): 563-578.

[15] Wenjie Xuan, Shaoli Huang, Juhua Liu, Bo Du. 2022. FCL-Net: Towards Accurate Edge Detection via Fine-scale Corrective Learning. *Neural Networks*, 2022, 145:248-259.

[16] Rodrigo Caye Daudt, Bertr Le Saux, and Alexandre Boulch. 2018. Fully convolutional siamese networks for change detection. *In 2018 25th IEEE International Conference on Image Processing (ICIP)*. IEEE, 4063–4067.

[17] Sheng Fang, Kaiyu Li, Jinyuan Shao, and Zhe Li. 2022. SNUNet-CD: A densely connected siamese network for change detection of VHR images. *IEEE Geoscience and Remote Sensing Letters* 19 (2022), 1–5.

[18] Hao Chen and Zhenwei Shi. 2020. A spatial-temporal attention-based method and a new dataset for remote sensing image change detection. *Remote Sensing* 12, 10 (2020), 1662.

[19] Jie Chen, Ziyang Yuan, Jian Peng, Li Chen, Haozhe Huang, Jiawei Zhu, Yu Liu, and Haifeng Li. 2020. DASNet: Dual attentive fully convolutional siamese networks for change detection in high-resolution satellite images. *IEEE Journal of Selected Topics in Applied Earth Observations and Remote Sensing* 14 (2020), 1194–1206.

[20] Hanhong Zheng, Maoguo Gong, Tongfei Liu, Fenlong Jiang, Tao Zhan, Di Lu, and Mingyang Zhang. 2022. HFA-Net: High frequency attention siamese network for building change detection in VHR remote sensing images. *Pattern Recognition*, 129: 108717.

[21] MA Lebedev, Yu V Vizilter, OV Vygolov, VA Knyaz, and A Yu Rubis. 2018. CHANGE DETECTION IN REMOTE SENSING IMAGES USING CONDITIONAL ADVERSARIAL NETWORKS. *International Archives of the Photogrammetry, Remote Sensing & Spatial Information Sciences* 42, 2 (2018).

[22] Shunping Ji, Shiqing Wei, and Meng Lu. 2019. Fully convolutional networks for multisource building extraction from an open aerial and satellite imagery data set. *IEEE Transactions on Geoscience and Remote Sensing* 57, 1 (2019), 574–586.

[23] Hui Zhang, Maoguo Gong, Puzhao Zhang, Linzhi Su, and Jiao Shi. 2016. Feature-level change detection using deep representation and feature change analysis for multispectral imagery. *IEEE Geoscience and Remote Sensing Letters* 13, 11 (2016), 1666–1670.

[24] Lichao Mou, Lorenzo Bruzzone, and Xiao Xiang Zhu. 2019. Learning spectral-spatial-temporal features via a recurrent convolutional neural network for change detection in multispectral imagery. *IEEE Transactions on Geoscience and Remote Sensing* 57, 2 (2019), 924–935.

[25] Yuli Sun, Lin Lei, Xiao Li, Hao Sun, and Gangyao Kuang. 2021. Nonlocal patch similarity based heterogeneous remote sensing change detection. *Pattern Recognition*, 109: 107598.

[26] Zongwei Zhou, Md Mahfuzur Rahman Siddiquee, Nima Tajbakhsh, and Jianming Liang. 2018. Unet++: A nested u-net architecture for medical image segmentation. In *Deep Learning in Medical Image Analysis and Multimodal Learning for Clinical Decision Support*. Springer, 3–11.

[27] Bo Fang, Li Pan, and Rong Kou. 2019. Dual learning-based siamese framework for change detection using bi-temporal VHR optical remote sensing images. *Remote Sensing* 11, 11 (2019), 1292.

[28] Rong Kou, Bo Fang, Gang Chen, and Lizhe Wang. 2020. Progressive domain adaptation for change detection using season-varying remote sensing images. *Remote Sensing* 12, 22 (2020), 3815.

[29] Jun-Yan Zhu, Taesung Park, Phillip Isola, and Alexei A Efros. 2017. Unpaired image-to-image translation using cycle-consistent adversarial networks. *In Proceedings of the IEEE International Conference on Computer Vision*. 2223–2232.

[30] Judy Hoffman, Eric Tzeng, Taesung Park, Jun-Yan Zhu, Phillip Isola, Kate Saenko, Alexei Efros, and Trevor Darrell. 2018. Cycada: Cycle-consistent adversarial domain adaptation. *In International Conference on Machine Learning*. PMLR, 1989–1998.

[31] Liangchen Song, Yonghao Xu, Lefei Zhang, Bo Du, Qian Zhang, and Xinggang Wang. 2020. Learning from synthetic images via active pseudo-labeling. *IEEE Transactions on Image Processing* 29 (2020), 6452–6465.

[32] Yaroslav Ganin and Victor Lempitsky. 2015. Unsupervised domain adaptation by backpropagation. *In International Conference on Machine Learning*. PMLR, 1180–1189.

[33] Eric Tzeng, Judy Hoffman, Kate Saenko, and Trevor Darrell. 2017. Adversarial discriminative domain adaptation. *In Proceedings of the IEEE Conference on Computer Vision and Pattern Recognition*. 7167–7176.

[34] Jingjing Li, Erpeng Chen, Zhengming Ding, Lei Zhu, Ke Lu, and Heng Tao Shen. 2021. Maximum density divergence for domain adaptation. *IEEE Transactions on Pattern Analysis and Machine Intelligence*, 43(11): 3918-3930.

[35] Jingjing Li, Mengmeng Jing, Hongzu Su, Ke Lu, Lei Zhu, and Heng Tao Shen. 2021. Faster domain adaptation networks. *IEEE Transactions on Knowledge and Data Engineering*, doi: 10.1109/TKDE.2021.3060473.

[36] Jingjing Li, Zhekai Du, Lei Zhu, Zhengming Ding, Ke Lu, and Heng Tao Shen. 2021. Divergence-agnostic Unsupervised Domain Adaptation by Adversarial Attacks. *IEEE Transactions on Pattern Analysis and Machine Intelligence*, doi: 10.1109/TPAMI.2021.3109287.

[37] Cheng Chen, Qi Dou, Hao Chen, Jing Qin, and Pheng-Ann Heng. 2019. Synergistic image and feature adaptation: Towards cross-modality domain adaptation for medical image segmentation. *In Proceedings of the AAAI Conference on Artificial Intelligence,* Vol. 33. 865–872.

[38] Yi-Hsuan Tsai, Wei-Chih Hung, Samuel Schulter, Kihyuk Sohn, Ming-Hsuan Yang, and Manmohan Chandraker. 2018. Learning to adapt structured output space for semantic segmentation. *In Proceedings of the IEEE Conference on Computer Vision and Pattern Recognition*. 7472–7481

[39] Tao Kong, Anbang Yao, Yurong Chen, and Fuchun Sun. 2016. Hypernet: Towards accurate region proposal generation and joint object detection. *In Proceedings of the IEEE Conference on Computer Vision and Pattern Recognition*. 845–853.

[40] Wei Liu, Dragomir Anguelov, Dumitru Erhan, Christian Szegedy, Scott Reed, Cheng-Yang Fu, and Alexander C Berg. 2016. Ssd: Single shot multibox detector. *In European Conference on Computer Vision.* Springer, 21–37. 47—49.